\pdfoutput=1

\documentclass[11pt]{article}

\usepackage[]{ACL2023}

\usepackage{times}
\usepackage{latexsym}
\usepackage{amsmath}
\usepackage{multirow}
\usepackage{bbm}
\usepackage{graphicx}
\usepackage{amssymb}
\usepackage{booktabs}
\usepackage{adjustbox}
\usepackage{xspace}
\usepackage{algpseudocode}
\usepackage{algorithm}
\usepackage{graphicx}
\usepackage{caption}
\usepackage{subcaption}
\usepackage{color, colortbl}
\usepackage{tablefootnote}
\usepackage[explicit]{titlesec}
\usepackage{todonotes}
\usepackage{enumitem}

\usepackage[T1]{fontenc}

\usepackage[utf8]{inputenc}

\usepackage{microtype}

\usepackage{inconsolata}

\title{Overcoming Catastrophic Forgetting in Massively Multilingual \\Continual Learning}

\author{Genta Indra Winata$^1$\thanks{\hspace{0.2cm}These authors contributed equally. $^\dagger$ The work was done while at Bloomberg.}\hspace{0.15cm}, Lingjue Xie$^{1*}$, Karthik Radhakrishnan$^{1*}$, Shijie Wu$^{1*}$, \\ \textbf{Xisen Jin}$^{2\dagger}$, 
\textbf{Pengxiang Cheng}$^{1}$, \textbf{Mayank Kulkarni}$^{3\dagger}$, \textbf{Daniel Preo\c{t}iuc-Pietro}$^{1}$ \\
  $^1$Bloomberg \quad $^2$University of Southern California \quad $^3$Amazon Alexa AI \\
  \texttt{\{gwinata,lxie91,kradhakris10,swu671,pcheng134,dpreotiucpie\}@bloomberg.net} \\
  \texttt{maykul@amazon.com, xisenjin@usc.edu}}

\begin{document}
\maketitle
\begin{abstract}
Real-life multilingual systems should be able to efficiently incorporate new languages as data distributions fed to the system evolve and shift over time. To do this, systems need to handle the issue of catastrophic forgetting, where the model performance drops for languages or tasks seen further in its past. In this paper, we study catastrophic forgetting, as well as methods to minimize this, in a massively multilingual continual learning framework involving up to 51 languages and covering both classification and sequence labeling tasks. We present \texttt{LR ADJUST}, a learning rate scheduling method that is simple, yet effective in preserving new information without strongly overwriting past knowledge. Furthermore, we show that this method is effective across multiple continual learning approaches. Finally, we provide further insights into the dynamics of catastrophic forgetting in this massively multilingual setup.

\end{abstract}

\section{Introduction}


Standard supervised NLP methods perform well when training on enough data from a uniform distribution. However, they fail to retain knowledge learnt in the past when sudden shifts occur in training data distributions. This effect of dropping performance on data from past distributions is commonly referred to as \textit{catastrophic forgetting}~\cite{mccloskey1989catastrophic,de2019episodic,biesialska2020continual}, where stability or preservation of knowledge is traded off for increased plasticity or the ability to acquire new knowledge. To tackle this issue, continual learning (CL) methods were proposed under various settings, such as limited compute or ability to store past data~\cite{lopez2017gradient,de2019episodic}. The data shifts commonly studied are obtained by training over a sequence of non-iid partitions~\cite{chaudhry2018efficient}, different tasks~\cite{jin-etal-2021-learn-continually}, or by training on various domains such as in task-oriented dialogue~\cite{madotto2021continual}, named entity recognition~\cite{monaikul2021continual}, part-of-speech~\cite{liu2020exploring}, and intent detection~\cite{wu2021pretrained}. 

\begin{figure}[!t]
    \centering
    \includegraphics[width=\linewidth]{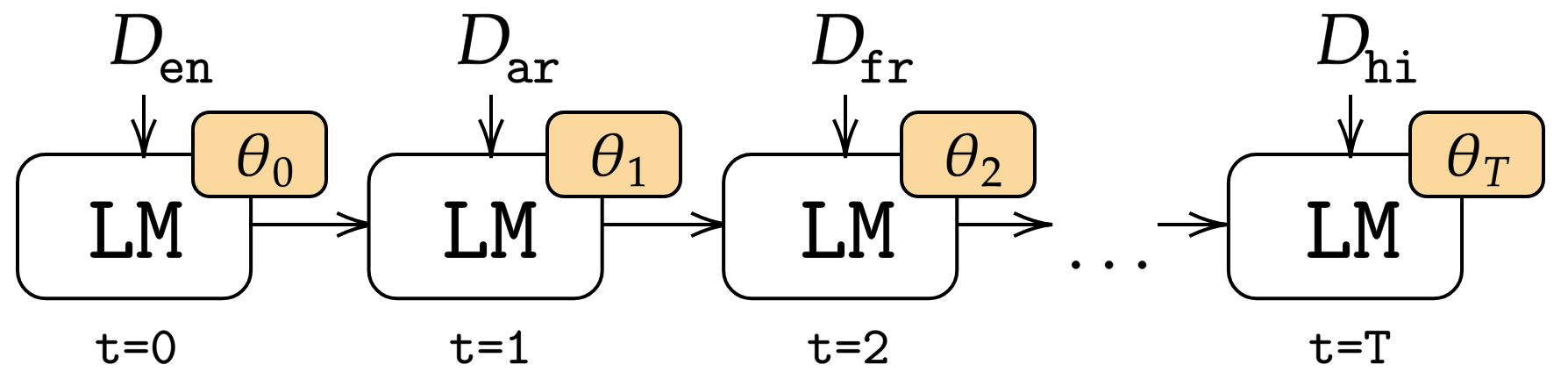}
    \caption{Multilingual Continual Learning: The model $\theta$ is trained sequentially with data from $T$ different languages.}
    \label{fig:diagram}
\end{figure}

Lifelong learning is key to the success of deployed multilingual systems, enabling the system to incorporate annotated data for new languages as they become available without costly retraining and redeployment of the entire system. This sequential availability of data for new languages is a common case of training data shift (see Figure~\ref{fig:diagram} for the task setup). Yet, the effect of catastrophic forgetting was not yet systematically studied for multi-lingual models with multiple diverse languages.~\citet{m2022cross} study continual learning in a cross-lingual setting limited to just six languages. The cross-lingual abilities of pre-trained models were found to drop when performing fine-tuning for a target language~\cite{liu2021preserving}, although applying continual learning approaches can effectively reduce the magnitude of the effect~\cite{lopez2017gradient}.

In this paper, we systematically study the effect of catastrophic forgetting and mitigation strategies in a massively multilingual setting covering up to 51 languages on three different tasks. We start by quantifying the extent to which forgetting happens when languages are presented to the model in sequence, identifying an up to 16\% F1 drop compared to training using all the data mixed.
Next, we propose \texttt{LR ADJUST}, a simple, yet effective, method to preserve the learned knowledge by adjusting the learning rate over time to alleviate the knowledge overwriting from the new language and preserve the previous learned knowledge. This method is orthogonal to continual learning methods and thus can be handily combined with any of these. We find that across three different CL methods, \texttt{LR ADJUST} helps further reduce the gap between a fully trained model and the CL setup. We conduct analysis on the aspect of cross-lingual transfer in backward and forward directions to measure the influence of the CL on previous tasks and its ability in zero-shot learning respectively.
Finally, we conduct analyses on the effects of catastrophic forgetting when first training on multiple languages jointly and when using a curriculum learning approach informed by language similarity.

\section{Massively Multilingual Continual Learning}

\subsection{Task Setup}
We define a curriculum of $T$ tasks as an ordered set of data sets $\mathcal{D} = \{D_1, D_2, ..., D_t, ..., D_T\}$ and model $\theta_t$, where $D_t$ is the data set with task $t$. In this case, the task is a distinct language. The weights of model $\theta$ are updated continuously $\theta_{t+1}$ $\leftarrow$ $f(\theta_{t}, D_t)$ by minimizing the log-likelihood over data set $D_t$ via gradient updates.



\subsection{Inter-task Learning Rate Adjustment}
\label{ssec:lradjust}
We propose \texttt{LR ADJUST}, a simple and effective method to adjust the learning rate when we start training on a new task. Our intuition is that models are susceptible to catastrophic forgetting when we provide a higher learning rate, so the learning rate should be toned down with time to ensure the preservation of the learned knowledge and to reduce the effect of overwriting the weights with the new knowledge. Learning rate adjustments have been studied in the context of incremental learning~\cite{cavalin2009evaluation,khreich2012survey} and for efficient optimization using schedules~\cite{ge2019step}. Concretely, the new learning rate is lowered every time as the following: $lr_{t} = max(lr_{min}, lr_{t-1} * \gamma$), with a weight $\gamma$, where $\gamma < 1$ and $lr_{min}$ is the minimum learning rate. The method is detailed in Algorithm~\ref{alg1}.

\begin{algorithm}[!t]
{\selectfont
\caption{Inter-task Learning Rate Adjustment (\texttt{LR ADJUST})}
\label{alg1}
\textbf{Require:} An ordered list of tasks $\mathcal{D} = \{D_1, D_2, ..., D_t, ..., D_T\}$ \\
\textbf{Require:} $\gamma$: learning rate adjustment coefficient, $lr_t$: learning rate at time $t$, $\theta$: model weights, $lr_{min}$: minimum learning rate
\begin{algorithmic}[1]
\State Randomly initialize the classifier on $\theta$
\For{\textbf{all} $D_t \in \mathcal{D}$}
  \State Adjust learning rate to $lr_{t} = max(lr_{min}, lr_{t-1} * \gamma$)
  \State Compute $\nabla_{\theta}\mathcal{L}_{D_t}(f_{\theta})$ using $D_t$
  \State $\theta_{t+1} \leftarrow \theta_t-lr_t \nabla_{\theta}\mathcal{L}_{D_t} (f_\theta)$  
\EndFor
\end{algorithmic}
}
\end{algorithm}

\subsection{Continual Learning Method}
We experiment with the following continual learning approaches:
\begin{itemize}[noitemsep,topsep=0pt,leftmargin=1em]
    \item \textbf{Experienced Replay}~\cite{de2019episodic} uses an episodic memory to store seen training data in memory and retrieve it from memory for fine-tuning. We schedule the replay step to be run every few iterations. During the replay step, we retrieve the data and fine-tune the model using the retrieved data. The number of stored data is constrained to ensure efficient memory use. And, we take a uniform distribution of samples across all labels.
    \item \textbf{Averaged GEM (A-GEM)}~\cite{chaudhry2018efficient} also utilizes an episodic memory $\mathcal{M}$ and is a more efficient implementation of GEM~\cite{lopez2017gradient} that computes the gradient constraints and minimizes the loss as follows:
    \begin{equation*}
        \mathcal{L}_t(\theta_t, \mathcal{M}) \leq \mathcal{L}_{t-1}(\theta_{t-1}, \mathcal{M}),
    \end{equation*}
    where loss $\mathcal{L}_t$ is constrained to be lower or equal to the loss $\mathcal{L}_{t-1}$.
    \item \textbf{Elastic Weight Consolidation (EWC)}~\cite{kirkpatrick2017overcoming} minimizes the following loss:
    \begin{equation*}
        \mathcal{L}_t(\theta) = \mathcal{L}_t(\theta) + \sum_i \frac{\lambda}{2}F_i (\theta_t - \theta^*_{t-1})^2,
    \end{equation*}
    where $F_i$ is the Fisher information matrix, $\lambda$ is a coefficient that sets how important the old task is compared to the new one, and $\theta^*_{t-1}$ is the previous learned weights. The $F_i$ is pre-computed after each task is completed and we incorporate the loss to the training on each gradient update.
\end{itemize}

\section{Experimental Setup}
\subsection{Data sets}

We use a multilingual natural language understanding data set, MASSIVE~\cite{fitzgerald2022massive} and a multilingual named entity recognition (NER) data set, WikiAnn~\cite{rahimi2019massively}. The MASSIVE data set consists of two tasks: intent classification and slot filling with 51 languages. The WikiAnn data set consists of 40 languages. We adopt the data splits from the original papers.

\subsection{Methods}

Our model architecture is an encoder-only multi-lingual model XLM-R$_{\text{BASE}}$~\cite{conneau2020unsupervised} with a classification layer for each task. All parameters in the model are updated during the training. The full hyper-parameters are listed in Appendix~\ref{sec:hyper-parameters}. The language order for the experiments are listed in Appendix~\ref{sec:language-order}. We experiment with the following approaches:
\begin{itemize}[noitemsep,topsep=0pt,leftmargin=1em]
    \item \texttt{MULTI}: A single multilingual model is trained on data mixed from all languages. This represents an upper bound to CL methods, as there are no memory or data sequencing constraints.
    \item \texttt{MONO}: A separate model is trained on all the supervised data for each language and applied to all inputs in that language.
    \item \texttt{VANILLA}: A single model is trained by sequentially presenting data from each language. The language order is selected randomly.
    \item CL Methods: We run \texttt{REPLAY}, \texttt{A-GEM}, and \texttt{EWC} to train a single model on data from each language presented sequentially.
    \item CL Methods + \texttt{LR ADJUST}: We run the CL methods with the learning rate adjustment method described in Section~\ref{ssec:lradjust}.
\end{itemize}

\subsection{Metrics}
We measure the ability of cross-lingual transfer using CL metrics adapted from~\citet{lopez2017gradient}. We define a matrix $R\in \mathbb{R}^{T \times T}$, where $R_{i,j}$ denotes the test score performance of the model on task $t_j$ when training the last sample from task $t_i$. We formally define the metrics as:

\subsubsection{Cross-lingual Forward Transfer (CFT)}
This metric represents the ability to perform zero-shot learning by evaluating on the test data from tasks/languages that are unseen in training. We formally define the metric as:
\begin{align*}
    CFT &= \frac{1}{T-1} \sum_{i=1}^{T-1} {\bar{X_{i}}}, \\
    \bar{X_{i}} &= \frac{1}{T-i} \sum_{j=i+1}^{T} {R_{i, j}},
\end{align*}
where $\bar{X_{i}}$ is the average performance of the languages that will be seen in the future ($t_{>i}$).

\subsubsection{Cross-lingual Backward Transfer (CBT)}
This metric measures the influence of learning a task $t_i$ on the performance of the previous tasks. We formally define the metric as the following:
\begin{equation*}
    CBT = \frac{1}{T-1} \sum_{i=1}^{T-1} {R_{T-1, i} - R_{i,i}}.
\end{equation*}

CBT practically measures the effect of catastrophic forgetting of past tasks after adding a new task to the model.

\section{Results and Analysis}

\begin{figure}[!t]
    \centering
    \includegraphics[width=\linewidth]{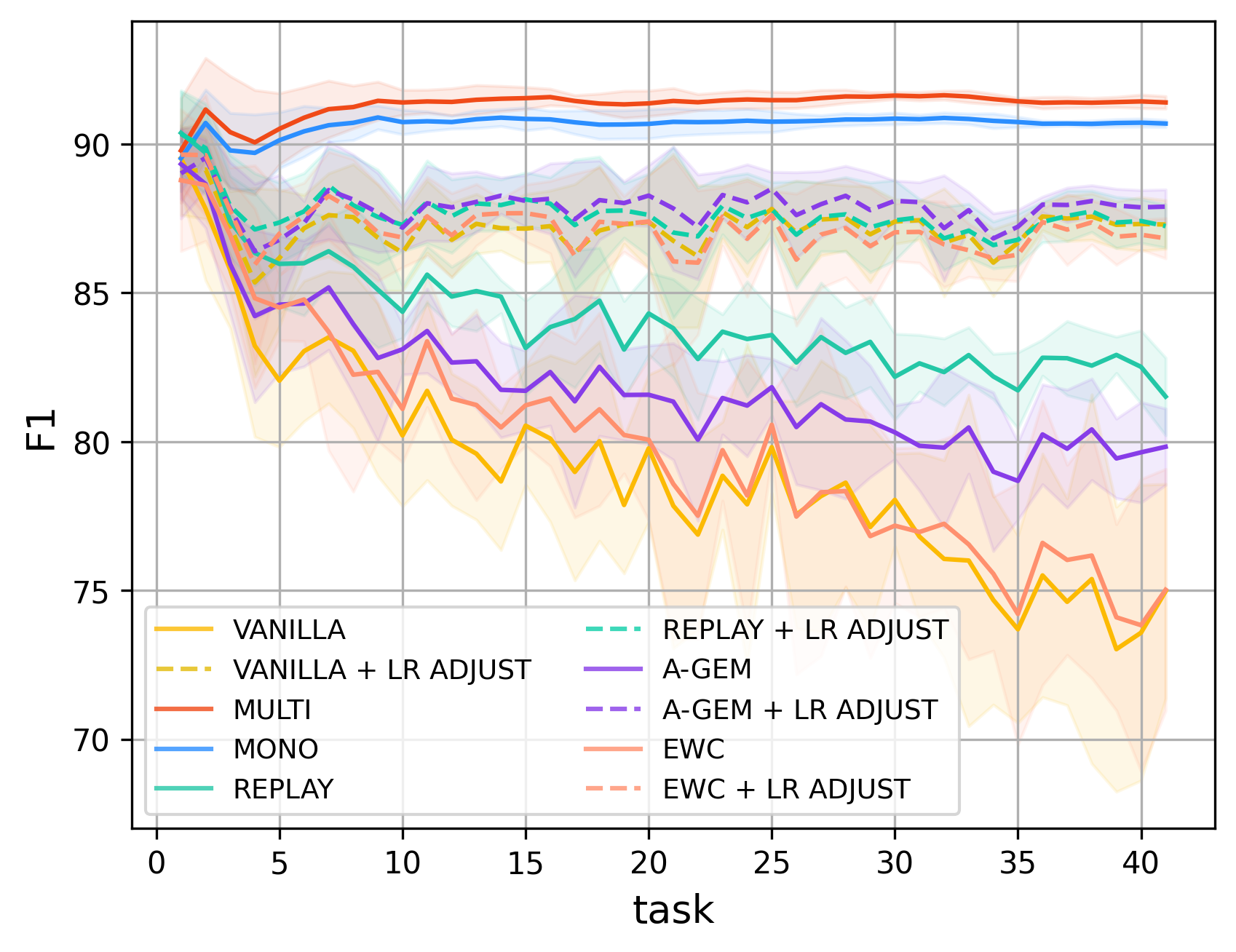}
    \caption{
    Average F1 scores and standard deviation over 5 runs on WikiAnn~\cite{rahimi2019massively} evaluated over increasing number of languages seen in training.}
    \label{fig:wikiann}

    \includegraphics[width=\linewidth]{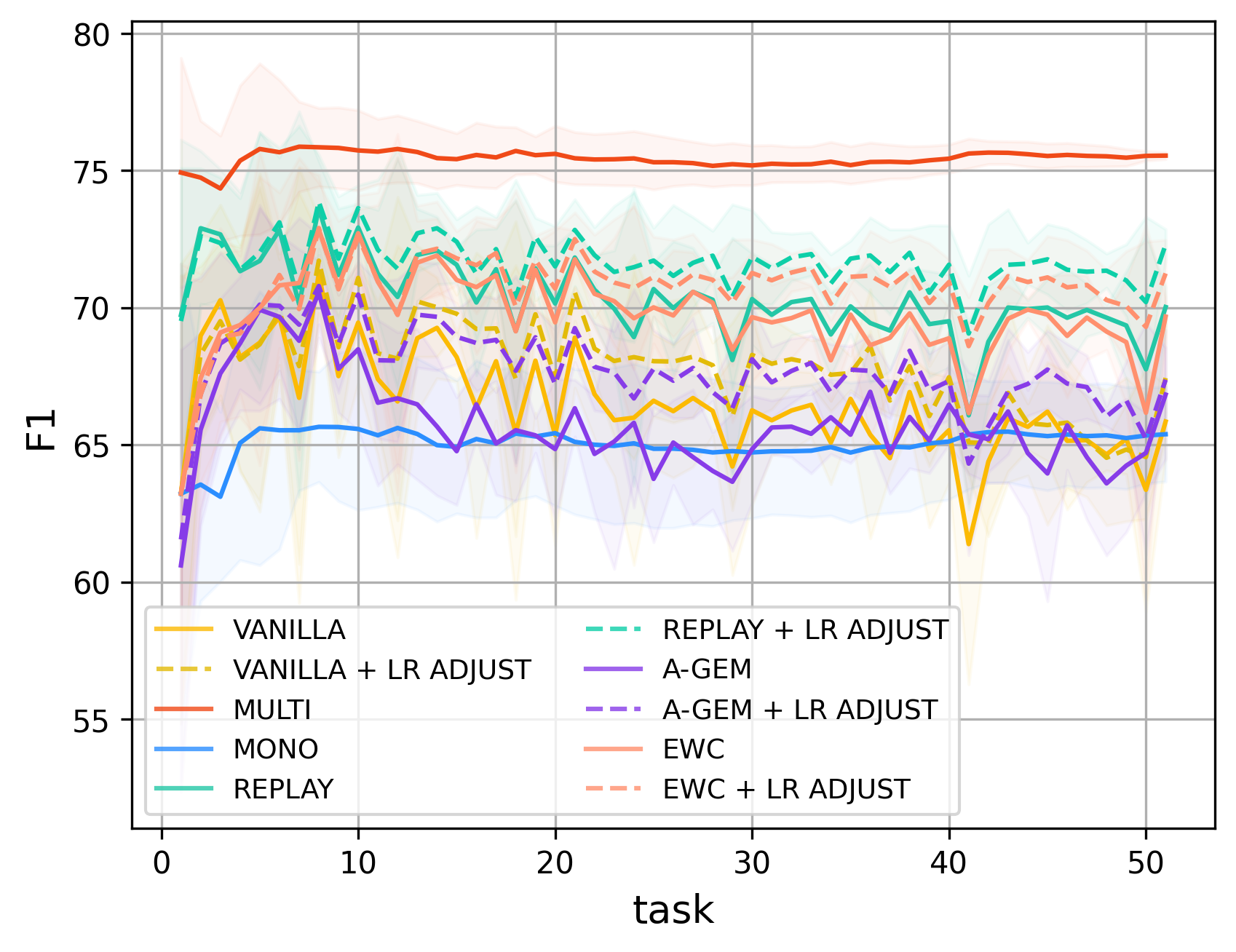}
    \caption{
    Average F1 scores and standard deviation over 5 runs on MASSIVE-Slot~\cite{fitzgerald2022massive} evaluated over increasing number of languages seen in training.}
    \label{fig:massive-slot}
\end{figure}


\begin{figure}[!ht]
    \centering
    \includegraphics[width=\linewidth]{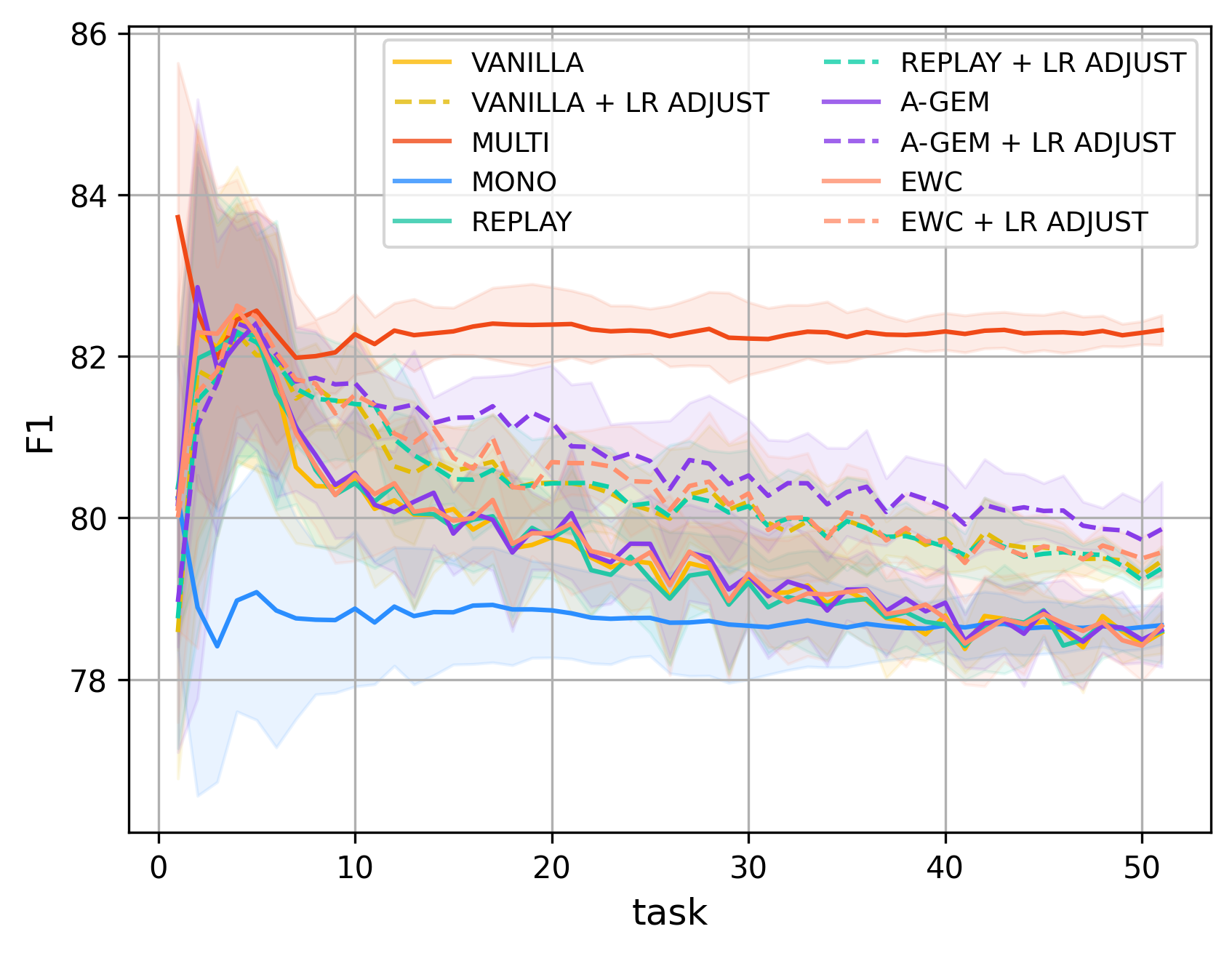}
    \caption{
    Average F1 scores and standard deviation over 5 runs on MASSIVE-Intent~\cite{fitzgerald2022massive} evaluated over increasing number of languages seen in training. }
    \label{fig:massive-intent}
\end{figure}

Figures~\ref{fig:wikiann},~\ref{fig:massive-slot} and \ref{fig:massive-intent} show performance numbers across the languages seen to that point in training for the three data sets. Each point on the graph shows the average and standard deviation of the F1 scores, obtained over 5 runs with different seeds. For the CL experiments, the seed also controls the order of languages in training. This can lead to higher variance across runs -- compared to mixing the same data across runs -- because the forgetting effect depends on the language order. Table~\ref{cross-lingual-transfer} shows the forward and backward transfer for the three data sets.

Results show the following:

\textbf{Catastrophic forgetting is present in multilingual continual learning}, with performance dropping quickly when training on languages sequentially (\texttt{VANILLA}). This method converges to between 4 -- 17 F1 lower than mixing the same data and training a full model every time (MULTI). We also see that this effect is present even when the performance of monolingual models is close to that of multilingual models, as in the case of WikiAnn. Training a full model on all languages (MULTI) always performs best, outperforming training on one language at a time (MONO) sometimes substantially (MASSIVE-Slot - 10 F1, MASSIVE-Intent - 4 F1), highlighting the importance of cross-lingual transfer and preserving information from past seen languages in a continual learning setup.

\textbf{Continual learning methods generally help} dampen the impact of forgetting. For example, in WikiAnn, the \texttt{REPLAY} and \texttt{A-GEM} CL methods reduce the backward transfer from 16.90 to 11.87 and 10.75 respectively, albeit \texttt{EWC} does not substantially improve the performance relative to the \texttt{VANILLA} method.

\textbf{\texttt{LR ADJUST}}, the learning rate adjustment scheme, \textbf{further reduces the gap} to the multi-task model significantly and consistently across languages when combined with any of the CL methods. For example, on the WikiAnn dataset, the backward transfer is reduced from 16.90 to just 3.59 and 3.79 for the \texttt{A-GEM} and \texttt{REPLAY} methods respectively, making multilingual CL feasible. Further, we see that using CL methods alone results in a continuous drop in performance as more languages are added, while adding \texttt{LR ADJUST} stabilizes average performance after the first few languages, resulting in a flatter curve.

Finally, we see that the \textbf{patterns of improvement hold when studying cross-lingual forward transfer}, which quantifies the zero-shot model performance on languages unseen in training. The continual learning approaches improve over sequential training (e.g.\ +4.45 on WikiAnn) and using \texttt{LR ADJUST} in addition further boosts performance (e.g.\ +9.25 for \texttt{VANILLA}, +5.83 for \texttt{REPLAY} on WikiAnn). This shows that the resulting models were able to retain essential and generalizable information for the task that is more universal across all languages.

\begin{table*}[!t]
\centering
\resizebox{0.77\linewidth}{!}{
\begin{tabular}{lcccccc}
\toprule
& \multicolumn{2}{c}{\textbf{WikiAnn}} & \multicolumn{2}{c}{\textbf{MASSIVE-Slot}} & \multicolumn{2}{c}{\textbf{MASSIVE-Intent}} \\
 & CFT  & CBT  & CFT  & CBT & CFT  & CBT \\ \midrule
VANILLA & 67.77 & -16.90 & 57.27 & -7.92 & 79.60 & -3.15\\
\quad+ LR ADJUST & 77.02 & -4.19 & 59.17 & -5.20 & 80.32 & -1.56 \\
EWC & 68.58 & -16.83 & 60.79 & -5.36 & 79.67 & -2.99 \\
\quad+ LR ADJUST & 76.78 & -4.50 & 62.30 & -3.36 & 80.46 & \textbf{-1.53} \\ 
A-GEM & 69.54 & -11.87 & 57.63 & -6.22 & 79.68 & -3.13 \\
\quad+ LR ADJUST & 77.26 & \textbf{-3.59} & 57.93 & \textbf{-2.48} & \textbf{80.73} & -1.65 \\ 
REPLAY & 72.22 & -10.75 & 60.38 & -6.04 & 79.23 & -3.18 \\ 
\quad+ LR ADJUST & \textbf{78.05} & -3.79 & \textbf{62.54} & -3.26 & 80.32 & -1.65 \\ \bottomrule
\end{tabular}
}
\caption{Cross-lingual Forward Transfer (CFT) and Cross-lingual Backward Transfer (CBT). Higher scores are better.}
\label{cross-lingual-transfer}
\end{table*}

\begin{figure}[!t]
    \centering
    \includegraphics[width=\linewidth]{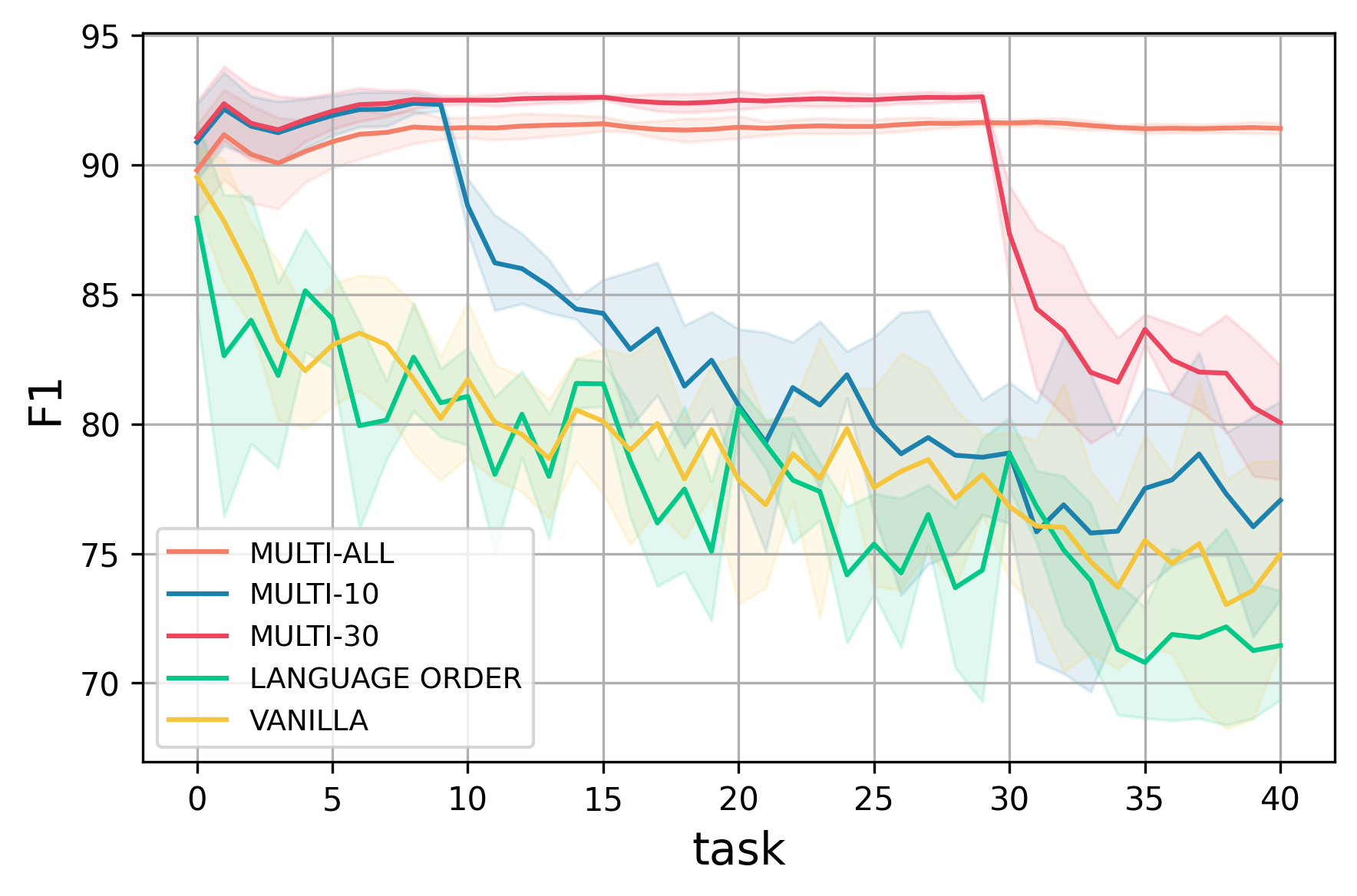}
    \caption{Multi-task strategies and training using the language order heuristics on WikiAnn. It shows the averaged F1 scores and standard deviation on 5 runs.}
    \label{fig:wikiann-mix}
\end{figure}

\subsection{Multi-task Training vs. Catastrophic Forgetting}
We conduct additional experiments to understand whether initially training on multiple languages at once can reduce the severity of catastrophic forgetting in the CL setup when new languages are added. We run a multi-task training on the first $k$ languages, where $k$ is 10 or 30, and then, we run the remaining languages sequentially on the WikiAnn data set. As shown in Figure~\ref{fig:wikiann-mix}, the model is more robust to forgetting when it is exposed to multi-task training with more languages with higher final average scores at the final task, but the graphs shows the performance still drops dramatically after being exposed to the first new language fed sequentially.

\subsection{The Role of Language Order}
To investigate the role of the language order on CL, we decide to reorder the language list by using heuristics. We start with two languages from the same family, as listed in Ethnologue~\cite{eberhard22simons}, and add all languages from the same family one by one, then switch to a new language family and continue the same process. We conjecture that seeing a similar language at an interval will allow a more effective cross-lingual transfer. Figure~\ref{fig:wikiann-mix} (\texttt{LANGUAGE ORDER}) displays the results, which indicate that performance does not improve after we manually select the languages, and its performance is similar to random ordering (\texttt{VANILLA}).

\section{Related Work}
CL aims to learn effectively over iterations by leveraging information from previously learned tasks~\cite{mccloskey1989catastrophic}. CL has been applied towards mitigating catastrophic forgetting in many computer vision tasks~\cite{lopez2017gradient}. Broadly speaking, CL methods can be classified \cite{biesialska-etal-2020-continual} into rehearsal (re-using training examples from prior task) \cite{rolnick2019experience,de2019episodic}, regularization (guide consolidation via additional loss) \cite{kirkpatrick2017overcoming}, memory \cite{lopez2017gradient, chaudhry2018efficient}, and architectural methods (using task-specific parameters) \cite{rusu2016progressive,madotto2021continual}.

\section{Conclusion}
We present the first study of catastrophic forgetting in a massively multilingual setting involving up to 51 languages on named entity recognition and natural language understanding tasks. We investigate continual learning methods and present a learning rate scheduling method that is simple yet effective in reducing the effects of catastrophic forgetting. Furthermore, we show that this method is effective across multiple continual learning methods. Finally, we provide analysis and further insights into the dynamics of catastrophic forgetting.

\section*{Acknowledgments}
We are grateful to Abhinav Singh and Shuyi Wang for feedback on a draft of this manuscript.

\section*{Limitations}
The experiment in this paper is limited to three common CL methods: Replay, A-GEM, and EWC. Due to compute resources, we experiment with XLM-R$_\text{BASE}$ and do not compare with other variants or larger base models. We acknowledge that the MASSIVE data set varies in coverage across language groups and evaluation could over-represent certain linguistic families~\cite{pikuliak-simko-2022-average}.

\section*{Ethics Statement}
In our experiments, we use publicly available data sets with permissive licenses for research experiments. We do not release new data or annotations as part of this work. There are no potential risks.

        


\bibliography{anthology,custom}
\bibliographystyle{acl_natbib}

\appendix

\section{Hyper-parameters}
\label{sec:hyper-parameters}

In all experiments, we run with five different seeds $\{42,52,62,72,82\}$ using a V100 32GB GPU and each run takes up to a week to finish.

\subsection{WikiAnn}
Table~\ref{hyper-parameters-wikiann} shows the hyper-parameters used in the experiments using WikiAnn dataset.

\begin{table}[!ht]
\centering
\resizebox{\linewidth}{!}{
\begin{tabular}{lccccccc}
\toprule
 & \textbf{MULTI} & \textbf{VANILLA} & \textbf{MONO} & \textbf{MULTI-10/30} & \textbf{REPLAY} & \textbf{A-GEM} & \textbf{EWC} \\ \midrule
LR & 1e-4 & 5e-5 & \begin{tabular}[c]{@{}c@{}}5e-5 (all except hi) \\ 1e-5 (hi)\end{tabular} & 5e-5 & - & - & - \\
LR Decay & 0.9997 & 0.999 & 0.9999 & 0.9997 & - & - & - \\
Batch Size & 32 & 32 & 32 & 32 & - & - & - \\
Max epochs & 20 & 20 & 20 & 20 & - & - & - \\
Early stopping & 5 & 5 & 5 & 5 & - & - \\
EWC\_Reg & - & - & - & - & - & - & 10000  \\
store\_memory\_prob & - & - & - & - & 0.0005 & 0.01 & - \\
max\_store\_num\_samples & - & - & - & - & 1E+7 & 1E+5 & - \\
retrieve\_num\_samples & - & - & - & - & 100 & - & - \\
run\_per\_step & - & - & - & - & 5000 & 2000 & - \\ \bottomrule
\end{tabular}
}
\caption{Hyper-parameters for WikiAnn data set.}
\label{hyper-parameters-wikiann}
\end{table}

\subsection{MASSIVE-Slot}
Table~\ref{hyper-parameters-massive-slot} shows the hyper-parameters used in the experiments using MASSIVE-Slot dataset.

\begin{table}[!ht]
\centering
\resizebox{\linewidth}{!}{
\begin{tabular}{lccccccc}
\toprule
 & \textbf{MULTI} & \textbf{VANILLA} & \textbf{MONO}  & \textbf{REPLAY} & \textbf{A-GEM} & \textbf{EWC} \\ \midrule
LR & 2e-5 & 1e-5 & 1e-5  & - & - & - \\
LR Decay & 0.9999 & 0.9999 & 0.9999  & - & - & - \\
Batch Size & 32 & 8 & 8 & - & - & - \\
Max epochs & 20 & 20 & 20 & 20 & - & - & - \\
Early stopping & 5 & 5 & 5 & 5 & - & - \\
EWC\_Reg & - & - & - & - & - &  10000  \\
store\_memory\_prob & - & - & - & 0.0005 & 0.01 & - \\
max\_store\_num\_samples & - & - & - & 1E+7 & 1E+5 & - \\
retrieve\_num\_samples & - & - & - & 100 & - & - \\
run\_per\_step & - & - & - & 5000 & 2000 & - \\ \bottomrule
\end{tabular}
}
\caption{Hyper-parameters for MASSIVE-Slot data set.}
\label{hyper-parameters-massive-slot}
\end{table}

\subsection{MASSIVE-Intent}
Additionally, we run the CL setup on MASSIVE-Intent dataset and the results are shown in Figure~\ref{fig:massive-intent}. Table~\ref{hyper-parameters-massive-intent} shows the hyper-parameters used in the experiments using MASSIVE-Intent dataset.

\begin{table}[!ht]
\centering
\resizebox{\linewidth}{!}{
\begin{tabular}{lccccccc}
\toprule
 & \textbf{MULTI} & \textbf{VANILLA} & \textbf{MONO} & \textbf{REPLAY} & \textbf{A-GEM} & \textbf{EWC} \\ \midrule
LR & 2e-5 & 2e-5 & 2e-5 & - & - & - \\
LR Decay & 0.999 & 0.995 & 0.995 & - & - & - \\
Batch Size & 32 & 32 & 32 & - & - & - \\
Max epochs & 20 & 20 & 20 & - & - & - \\
Early stopping & 5 & 5 & 5 & - & - & - \\
EWC\_Reg & - & - & - & - & - & 10000  \\
store\_memory\_prob & - & - & - & 0.0005 & 0.01 & - \\
max\_store\_num\_samples & - & - & - & 1E+7 & 1E+5 & - \\
retrieve\_num\_samples & - & - & - & 100 & - & - \\
run\_per\_step & - & - & - & 5000 & 2000 & - \\ \bottomrule
\end{tabular}
}
\caption{Hyper-parameters for MASSIVE-Intent data set.}
\label{hyper-parameters-massive-intent}
\end{table}

\section{Language Order}
\label{sec:language-order}

We randomly shuffle the language order for each seed. Tables~\ref{language-order-wikiann} and~\ref{language-order-massive} show the language order we use in the experiments for WikiANN and MASSIVE datasets, respectively.

\begin{table}[!ht]
\centering
\resizebox{0.75\linewidth}{!}{
\begin{tabular}{lc}
\toprule
\textbf{seed} & \textbf{languages} \\ \midrule
42 & el, bn, en, ta, ms, mk, ro, es, \\ & bs, sk, it, pl, lv, hr, et, sq, \\ & sv, nl, fa, lt, id, ru, tl, pt, \\ & hu, he, uk, sl, bg, af, tr, no, \\ & ca, cs, de, fi, fr, hi, ar, da, vi \\
52 & el, sl, hr, he, fa, it, lt, tl, \\ & mk, cs, pl, hu, bs, tr, uk, fr, \\ & ta, pt, sq, da, ms, no, et, vi, ar, \\ & af, id, fi, es, ca, ru, sv, en, \\ &  de, bg, nl, lv, ro, sk, bn, hi \\
62 & et, fi, ar, sv, ms, fa, sq, tr, \\ & it, ru, no, el, da, pl, hi, bg, \\ & cs, nl, hr, sl, mk, he, lv, tl, vi, \\ & bn, ro, id, de, af, ca, uk, sk, \\ & en, lt, hu, pt, fr, bs, es, ta \\
72 & fr, uk, mk, hr, ar, sl, sk, ta, \\ & bn, hi, ca, ro, pt, cs, fa, nl, en, \\ & he, pl, el, bg, sv, no, ru, da, ms, \\ & tl, af, id, vi, et, fi, it, de, \\ & hu, lv, sq, lt, es, tr, bs \\
82 & de, fi, ar, pl, pt, da, ms, hu, \\ & et, lv, ca, lt, af, fa, sq, mk, id, \\ & it, ta, sl, tr, ro, uk, bs, hi, vi, \\ & cs, bn, nl, tl, fr, no, bg, sv, \\ & he, en, es, hr, sk, ru, el \\ \bottomrule
\end{tabular}
}
\caption{Language Order for Experiments with WikiAnn.}
\label{language-order-wikiann}
\end{table}

\begin{table}[!ht]
\centering
\resizebox{\linewidth}{!}{
\begin{tabular}{lc}
\toprule
\textbf{seed} & \textbf{languages} \\ \midrule
42 & kn-IN, ka-GE, is-IS, fa-IR, bn-BD, \\ & tl-PH, ko-KR, en-US, mn-MN, hu-HU, \\ & my-MM, ja-JP, fi-FI, az-AZ, sq-AL, \\ & sl-SL, es-ES, km-KH, pt-PT, af-ZA, \\ & te-IN, id-ID, nl-NL, zh-CN, sw-KE, \\ & ms-MY, ml-IN, it-IT, jv-ID, ta-IN, \\ & tr-TR, ro-RO, nb-NO, th-TH, fr-FR, \\ & zh-TW, vi-VN, ar-SA, lv-LV, ru-RU, \\ & cy-GB, pl-PL, da-DK, el-GR, he-IL, \\ & hi-IN, hy-AM, ur-PK, am-ET, de-DE, sv-SE \\
52 &[nl-NL, is-IS, bn-BD, id-ID, en-US, 
\\ & my-MM, kn-IN, he-IL, ja-JP, da-DK, \\ & sq-AL, hu-HU, tl-PH, lv-LV, sw-KE, \\ & zh-TW, mn-MN, fi-FI, am-ET, zh-CN, \\ & fr-FR, sl-SL, sv-SE, ta-IN, it-IT, \\ & vi-VN, hi-IN, ur-PK, cy-GB, pt-PT, \\ & de-DE, ro-RO, ru-RU, km-KH, pl-PL, \\ & te-IN, af-ZA, ml-IN, jv-ID, fa-IR, \\ & th-TH, es-ES, el-GR, ar-SA, ko-KR, \\ & ka-GE, ms-MY, nb-NO, tr-TR, az-AZ, hy-AM \\
62 & sv-SE, az-AZ, ko-KR, ja-JP, el-GR, \\ & ru-RU, my-MM, ka-GE, ur-PK, vi-VN, \\ & tl-PH, pt-PT, fr-FR, kn-IN, tr-TR, \\ & en-US, fi-FI, sl-SL, he-IL, hy-AM, \\ & ml-IN, ar-SA, sw-KE, da-DK, te-IN, \\ & cy-GB, it-IT, id-ID, zh-TW, lv-LV, \\ & km-KH, pl-PL, nl-NL, ms-MY, am-ET, \\ & de-DE, sq-AL, hu-HU, af-ZA, th-TH, \\ & zh-CN, nb-NO, es-ES, jv-ID, ta-IN, \\ & is-IS, mn-MN, hi-IN, bn-BD, fa-IR, ro-RO \\
72 & fi-FI, tl-PH, tr-TR, da-DK, zh-TW, \\ & hi-IN, sw-KE, ko-KR, ms-MY, lv-LV, \\ & cy-GB, az-AZ, ml-IN, kn-IN, sv-SE, \\ & hy-AM, de-DE, id-ID, vi-VN, it-IT, \\ & te-IN, fr-FR, my-MM, ta-IN, es-ES, \\ & hu-HU, nb-NO, pt-PT, ro-RO, ar-SA, \\ & nl-NL, af-ZA, mn-MN, ru-RU, am-ET, \\ & en-US, km-KH, he-IL, ja-JP, el-GR, \\ & zh-CN, is-IS, ka-GE, sq-AL, pl-PL, \\ & th-TH, jv-ID, fa-IR, ur-PK, sl-SL, bn-BD \\
82 & az-AZ, he-IL, am-ET, fr-FR, ta-IN, \\ & ka-GE, ja-JP, hy-AM, bn-BD, ml-IN, \\ & ro-RO, pl-PL, jv-ID, pt-PT, nl-NL, \\ & tr-TR, mn-MN, zh-TW, ko-KR, ur-PK, \\ & af-ZA, cy-GB, sq-AL, vi-VN, hi-IN, \\ & km-KH, tl-PH, kn-IN, sw-KE, it-IT, \\ & sv-SE, sl-SL, de-DE, el-GR, is-IS, \\ & fi-FI, da-DK, ru-RU, ms-MY, lv-LV, \\ & ar-SA, th-TH, hu-HU, te-IN, es-ES, \\ & fa-IR, id-ID, nb-NO, my-MM, zh-CN, en-US \\ \bottomrule
\end{tabular}
}
\caption{Language Order for Experiments with MASSIVE.}
\label{language-order-massive}
\end{table}

\section{Geographical Information of Languages}
Table~\ref{language-geo} shows all languages' language families and subgroups on NusaX and MASSIVE datasets.

\begin{table}[!ht]
\centering
\resizebox{\linewidth}{!}{
\begin{tabular}{lccc}
\toprule
\textbf{Language Code} & \textbf{Name} & \textbf{Family} & \textbf{Subgroup} \\ \midrule
af / af-ZA & Afrikaans & Indo-European & Germanic \\
am-ET & Amharic & Afro-Asiatic & Semitic \\
ar / ar-SA & Arabic & Afro-Asiatic & Semitic \\
az-AZ & Azerbaijani & Turkic & Southern \\
bn / bn-BD & Bengali & Indo-European & Indo-Iranian \\
bg & Bulgarian & Indo-European & Balto-Slavic \\
bs & Bosnian & Indo-European & Balto-Slavic \\
ca & Catalan & Indo-European & Italic \\
cs & Czech & Indo-European & Balto-Slavic \\
cy-GB & Welsh & Indo-European & Celtic \\
da / da-DK & Danish & Indo-European & Germanic \\
de / de-DE & German & Indo-European & Germanic \\
en / en-US & English & Indo-European & Germanic \\
el / el-GR & Greek & Indo-European & Greek \\
es / es-ES & Spanish & Indo-European & Italic \\
et & Estonian & Uralic & Finnic \\
fa & Persian & Indo-European & Indo-Iranian \\
fi / fi-FI & Finnish & Uralic & Finnic \\
fr / fr-FR & French & Indo-European & Italic \\
id / id-ID & Indonesian & Austronesian & Malayo-Polynesian \\
is-IS & Icelandic & Indo-European & Germanic \\
it / it-IT & Italian & Indo-European & Italic \\
ja-JP & Japanese & Japonic & Japanese \\
jv-ID & Javanese & Austronesian & Malayo-Polynesian \\
he / he-IL & Hebrew & Afro-Asiatic & Semitic \\
hi / hi-IN & Hindi & Indo-European & Indo-Iranian \\
hr & Croatian & Indo-European & Balto-Slavic \\
hu / hu-HU & Hungarian & Uralic & Hungarian \\
hy-AM & Armenian & Indo-European & Armenian \\
ka-GE & Georgian & Kartvelian & Georgian \\
km-KH & Khmer & Austro-Asiatic & Mon-Khmer \\
kn-IN & Kannada & Dravidian & Southern \\
ko-KR & Korean & Koreanic & Korean \\
lt & Lithuanian & Indo-European & Balto-Slavic \\
lv / lv-LV & Latvian & Indo-European & Balto-Slavic \\
ml-IN & Malayalam & Dravidian & Southern \\
mn-MN & Mongolian & Mongolic & Eastern \\
ms / ms-MY & Malay & Austronesian & Malayo-Polynesian \\
mk & Macedonian & Indo-European & Balto-Slavic \\
my-MM & Burmese & Sino-Tibetan & Tibeto-Burman \\
nb-NO & Norwegian & Indo-European & Germanic \\
nl / nl-NL & Dutch & Indo-European & Germanic \\
no & Norwegian & Indo-European & Germanic \\
pl / pl-PL & Polish & Indo-European & Balto-Slavic \\
pt / pt-PT & Portuguese & Indo-European & Italic \\
ro / ro-RO & Romanian & Indo-European & Italic \\
ru / ru-RU & Russian & Indo-European & Balto-Slavic  \\
sl / sl-SL & Slovenian & Indo-European & Balto-Slavic \\
sk & Slovak & Indo-European & Balto-Slavic \\
sq / sq-AL & Albanian & Indo-European & Albanian \\
sw-KE & Swahili & Niger-Congo & Atlantic-Congo \\
sv / sv-SE & Swedish & Indo-European & Germanic \\
ta / ta-IN & Tamil & Dravidian & Southern \\
te-IN & Telugu & Dravidian & South-Central \\
th-TH & Thai & Kra-Dai & Kam-Tai \\
tl / tl-PH & Tagalog & Austronesian & Malayo-Polynesian \\
tr / tr-TR & Turkish & Turkic & Southern \\
ur-PK & Urdu & Indo-European & Indo-Iranian  \\
uk & Ukrainian & Indo-European & Balto-Slavic \\
vi / vi-VN & Vietnamese & Austro-Asiatic  & Mon-Khmer \\ 
zh-CN / zh-TW & Chinese & Sino-Tibetan  & Chinese \\ \bottomrule
\end{tabular}
}
\caption{Geographical information of languages under study. The language family is based on Ethnologue~\cite{david2019ethnologue}.}
\label{language-geo}
\end{table}



\end{document}